\def\year{2019}\relax
\documentclass[letterpaper]{article} 
\usepackage{aaai19}  
\usepackage{times}  
\usepackage{helvet}  
\usepackage{courier}  
\usepackage{url}  
\usepackage{graphicx}  

\usepackage{subfigure}
\usepackage{booktabs}
\usepackage{amsmath}
\usepackage{color}
\usepackage{algorithm,algorithmicx,algpseudocode}

\usepackage{latexsym}
\usepackage{breakcites}
\usepackage{url}
\usepackage{multirow}
\usepackage{latexsym}
\usepackage{mathrsfs}
\usepackage{amsfonts}
\usepackage{amssymb}
\usepackage{amsmath}
\usepackage{dsfont}
\usepackage{bm}
\usepackage{url}
\usepackage{lipsum}
\usepackage{titlesec}
\usepackage{times}
\usepackage{graphicx} 
\usepackage{subfigure}

\usepackage{tikz-dependency}
\usepackage{pgfplots}
\usepackage{xcolor}
\usepackage{array}
\usepackage{enumitem}
\usepackage{booktabs}
\usepackage{colortbl}

\let\OldStatex\Statex

\def\W{\mathbf{W}}

\def\bb{\mathbf{b}}

\def\bh{\mathbf{h}}

\def\h{\mathbf{h}}

\def\w{\mathbf{w}}
\def\W{\mathbf{W}}

\def\bb{\mathbf{b}}

\def\h{\mathbf{h}}

\def\bh{\mathbf{h}}

\def\bg{\mathbf{g}}
\def\bb{\mathbf{b}}
\def\bW{\mathbf{W}}

\def\bv{\mathbf{v}}
\def\bu{\mathbf{u}}

\renewcommand{\Statex}[1][3]{%
  \setlength\@tempdima{\algorithmicindent}%
  \OldStatex\hskip\dimexpr#1\@tempdima\relax}
  
\newcommand{\tabincell}[2]{\begin{tabular}{@{}#1@{}}#2\end{tabular}}
\newcommand\newcite[1]{\citeauthor{#1} [\citeyear{#1}]}
\DeclareMathOperator*{\Softmax}{Softmax}
\DeclareMathOperator*{\softmax}{Softmax}
\DeclareMathOperator*{\CRF}{CRF}

\algnewcommand\LeftComment[2]{%
\hspace{#1\algindent}$\triangleright$ \eqparbox{COMMENT}{#2} \hfill %
}

\newenvironment{itemize*}%
 {\begin{itemize}%
  \setlength{\itemsep}{0pt}%
  \setlength{\parskip}{0pt}}%
 {\end{itemize}}
 \newenvironment{enumerate*}%
 {\begin{enumerate}%
  \setlength{\itemsep}{0pt}%
  \setlength{\parskip}{0pt}}%
 {\end{enumerate}}

\def\W{\mathbf{W}}

\def\bb{\mathbf{b}}

\def\bh{\mathbf{h}}

\def\h{\mathbf{h}}

\def\w{\mathbf{w}}
\def\W{\mathbf{W}}

\def\bb{\mathbf{b}}

\def\h{\mathbf{h}}

\def\bh{\mathbf{h}}

\def\bg{\mathbf{g}}
\def\bb{\mathbf{b}}
\def\bW{\mathbf{W}}

\def\bv{\mathbf{v}}
\def\bu{\mathbf{u}}

\let\OldStatex\Statex
\renewcommand{\Statex}[1][3]{%
  \setlength\@tempdima{\algorithmicindent}%
  \OldStatex\hskip\dimexpr#1\@tempdima\relax}

\makeatletter

\author{Pengfei Liu$\dag \ddag$, Shuaichen Chang, Xuanjing Huang$\dag$, Jian Tang$\ddag$, Jackie Chi Kit Cheung$\ddag \sharp$\\
 $\dag$School of Computer Science, Fudan University, Shanghai Insitute of Intelligent Electroics \& Systems \\
    $\ddag$MILA  \& $\sharp$McGill University  \&  The Ohio State University \\
\{pfliu14,xjhuang\}@fudan.edu.cn, chang.1692@osu.edu,jian.tang@hec.ca,jcheung@cs.mcgill.ca}

\title{Contextualized Non-local Neural Networks for Sequence Learning}

\date{}

\begin{document}
\maketitle
\begin{abstract}
Recently, a large number of neural mechanisms and models have been proposed for sequence learning, of which self-attention, as exemplified by the Transformer model, and graph neural networks (GNNs) have attracted much attention. In this paper, we propose an approach that combines and draws on the complementary strengths of these two methods.
Specifically, we propose contextualized non-local neural networks (CN$^{\textbf{3}}$), which can both dynamically construct a task-specific structure of a sentence and leverage rich local dependencies within a particular neighbourhood.

Experimental results on ten NLP tasks in text classification, semantic matching, and sequence labelling show that our proposed model outperforms competitive baselines and discovers task-specific dependency structures, thus providing better interpretability to users.
\end{abstract}

\section{Introduction}

Learning the representation of sequences is a fundamental task, which requires deep understanding
of both the \textbf{complex structure} of sentences \cite{biber1998corpus} and the
\textbf{contextualized representation} of words \cite{peters2018deep}.
Recent successful approaches replace the classical compositional functions in neural networks (i.e. CNNs or RNNs)
with mechanisms based on self-attention, of which the most effective model is the \textit{Transformer}, having achieved state-of-the-art performance on machine translation \cite{vaswani2017attention} and parsing \cite{kitaev2018constituency}.
The success of Transformer can be attributed to its non-local structure bias, in which dependencies between any pair of words can be modelled \cite{baerman2015oxford,wang2017non}. This property allows it to dynamically learn both the syntactic and semantic structures of sentences \cite{vaswani2017attention}.
Despite its success, the lack of local bias (local dependencies that exist over adjacent words \cite{baerman2015oxford,futrell2017memory}) limits its capacity for learning contextualized representations of words\footnote{Our experiments also show that Transformer performs worse than the models with local bias, especially on sequence labeling tasks.}.

\begin{table}[!t]
\small

\centering
\begin{tabular}{llccc}
\toprule

\textbf{Challenges} & \textbf{Bias} & $\mathbf{G.}$ & $\mathbf{T.}$ & $\mathbf{Ours}$\\
\midrule
Com-Structure                & \tabincell{l}{Non-local}       & $\times$  & $\checkmark$  & $\checkmark$\\
Con-Representation           & \tabincell{l}{Local}            & $\checkmark$      & $\times$ & $\checkmark$\\
\bottomrule
\end{tabular}
\caption{
Com-Structure denotes ``\textbf{complicated structures of sentences}'' while Con-Representation denotes ``\textbf{contextualized representations of words}''.
G. and T. represent graph neural network and Transformer respectively.
The first two columns shows main challenges for sentence learning and corresponding required  bias:
\textbf{Non-local bias}: dependencies can be built with any neighbor \cite{baerman2015oxford,wang2017non}.
\textbf{Local bias}: local dependencies exist over adjacent word \cite{baerman2015oxford,futrell2017memory}.
The last three columns list comparison of typical sequence learning approaches for incorporating local and non-local bias.
} \label{tab:property}
\end{table}

In contrast to Transformer, another line of work aims to model sequences with \textit{graph neural networks} (GNNs).
While GNNs are capable of learning local contextual information flexibly by encoding attribute features \cite{battaglia2018relational}, it is less clear how to
effectively utilize GNNs for sequence learning, since there is no single structural representation of a sentence that is well suited to all tasks.

A common choice is to use the syntactic dependencies between words in a sentence to determine the sentential graph structure, which reduces the model to a tree-structured neural network \cite{tai2015improved}.
Inspired by Transformer \cite{vaswani2017attention}, we aim to improve upon this fixed structure by learning a task-dependent graphical representation, which should better capture the dependencies that matter for the end task.

In this paper, we draw from both lines of research and propose a model which can benefit from their complementary strengths.
On the one hand, the success of Transformer motivates us to explore dynamic ``\textsc{graph}'' construction.
Rather than defining a hard-coded graph for the sentence, we incorporate non-local bias into GNNs, learning the sentence structure dynamically on different tasks.
On the other hand, the advantage of the graph neural network framework itself is to provide rich local information by encoding node or edge attributes, which can make up for the deficiency in Transformer.

Consequently, we propose a contextualized non-local networks by extending self-attention (Transformer) to GNNs, which support highly flexible representations in two ways. First, the representations of attributes (feature encodings of nodes and edges); second, the structure of the graph itself (dynamic learning of task-dependent structures).
Notably, these two advantages of the proposed model enable us to better learn the representation of sequences in terms of contextualized representation of words and complicated structures of sentences.

We conduct extensive experiments on ten sequence learning tasks, from text classification and semantic matching, to sequence labelling. Experimental results show that our proposed approach outperforms competitive baselines or achieves comparable results in all tasks. Additionally, we are able to discover task-dependent structures among words and provide better interpretablility to users.

\begin{table*}[!t]
\center \footnotesize
\tabcolsep0.07in
\begin{tabular}{p{3.8cm}{l}{l}{l}{l}}
\toprule

\center \multirow{2}{*}{ \textbf{Models}}  & \multicolumn{2}{c}{\textbf{Interaction}} & \multirow{2}{*}{\textbf{Locality Bias}} & \multirow{2}{*}{\textbf{Eligible Tasks}}\\
\cmidrule(lr){2-3}  
 &  \textbf{Scope} & \textbf{Functions} \\

\midrule
 BOW \\\cite{fan2008liblinear}               &  None              &  None                                               & Local    & \texttt{T-Sent} \& \texttt{T-Word}   \\
 \midrule

S-Graph \\\cite{mihalcea2004textrank}           & \tabincell{l}{All pairs of \\words}             &  $PMI(\w_i,\w_j)$                                     & \tabincell{l}{Non-Local} & \texttt{T-Sent} \\
\midrule

CNN \\\cite{kalchbrenner2014convolutional}               &	 Fixed window  & $f_{\theta}(\w_i) * g_{\theta}(\w_j)$               & Local & \texttt{T-Sent}   \\
\midrule
SeqRNN \\\cite{liu2015multitimescale}           &	Adjacent words    & $f_{\theta}(\w_i,\w_j)$                             & Local  & \texttt{T-Sent}   \\
\midrule
TreeRNN \\ \cite{tai2015improved,zhu2015long,liu2017dynamic}           &	Child words       &  $f_{\theta}(\w_i,\w_j)$                            & \tabincell{l}{Local}  & \texttt{T-Sent}  \\
\midrule
GraphCNN \\ \cite{marcheggiani2017encoding}           &	Child words       &  $f_{\theta}(\w_i,\w_j)$                            & \tabincell{l}{Local}  & \texttt{T-Word} \\
\midrule
SelfAtt1 \\ \cite{yang2016hierarchical,lin2017structured}         &	 Adjacent words   &  $f_{\theta}(\w_i,\w_j)$                            & Local  & \texttt{T-Sent} \\
\midrule
SelfAtt2 \\\cite{cheng2016long,7}       &	 Preceding words    &  $\sum_{i\in \Phi} \alpha_i \cdot  f_{\theta}(\w_i)$  & Local  & \texttt{T-Sent} \& \texttt{T-Word}  \\
\midrule
SelfAtt3 \\\cite{vaswani2017attention}         &	 \tabincell{l}{All pairs of \\words}            & $\sum_{i\in \Phi} \alpha_i \cdot [\w_i,\mathbf{p_i}]$  & \tabincell{l}{Non-local}  & \texttt{T-Word} \\
\midrule
\rowcolor[gray]{.7}
{Our Work}          &	 \tabincell{l}{All pairs of \\words}            & $\sum_{i\in \Phi} \alpha_i \cdot f_{\theta}(\w_i)$  & \tabincell{l}{Contextualized Non-local}  & \texttt{T-Sent} \& \texttt{T-Word}  \\
\bottomrule
\end{tabular}
\caption{
A comparison of published  approaches for sentence representations.
S-Graph denotes statistical graph.
$f(\cdot)$ and $g(\cdot)$ are parameterized functions. $\w$ denotes the representation of word and $\mathbf{p}$ is a vector relating to positional information. $\alpha_i$ is a scalar and $\Phi$ is a set of indexes for those words within a interaction scope.
The \textbf{2nd column} shows which part of words can interact with each other while the
\textbf{3rd column} shows how to compose these words.
The \textbf{4th column} lists what kinds of knowledge the interaction methods depend on.
The \textbf{last column} represents which tasks can be processed by corresponding models. ``\texttt{T-Sent}'' denotes those tasks who require sentence-level representations, such as text classification, while ``\texttt{T-word}'' represent tasks which rely on word-level representations, such as  sequence labelling, machine translation.
``-'': S-Graph usually calls for complicated ranking algorithm and can not directly obtain a real-valued vector to represent the sentence meanings.
``*'' SelfAtt3 is proposed for machine translation, rather than sentence level representation learning.
} \label{tab:all-models}
\end{table*}

We summarize our contributions as below:
\begin{enumerate}
\item We analyze two challenges for sequence learning (structure of sentences and contextualized word representations) \textbf{from the perspective of models' locality biases}, which encourages us to find the complementarity between Transformer and graph neural networks.
\item We draw on the complementary strengths of Transformer and GNNs, proposing a contextualized non-local neural network (CN$^\textbf{3}$), in which structures of sentences can be learned on the fly and words in sentence can be contextualized flexibly.
\item We perform a comprehensive analysis of existing sequence learning approaches in a unified way and conduct extensive experiments on a number of natural language understanding tasks. Experimental results show that our proposed models achieve comparable results in these tasks as well as offer better interpretability to users.
\end{enumerate}

\section{Related Work} \label{relate}

In this paper, we propose to utilize the \textbf{locality bias} to systematically analyze existing sequence learning models in a unified way.
Here, the locality bias is one kind of inductive bias that local dependencies exist over adjacent words, which is first well established in phonology \cite{van2005domain} and then also be explored in natural language \cite{baerman2015oxford,futrell2017memory}.
Typically, local and non-local bias are two common locality biases implicitly existing in different models for sequence learning.

Next, we will elaborate on explanations following related lines and as shown in Table~\ref{tab:all-models}, we present a summary of existing models by
highlighting differences among interaction methods for different words, locality bias, as well as the eligible tasks that corresponding methods can handle.

\paragraph{Neural network-based Sequence Modelling.}
Neural networks provide an effective way to model the dependencies between different words via {parameterized composition functions}.
Some examples of composition functions involve recurrent neural networks with long short-term memory (LSTM) \cite{liu2015multitimescale}, convolutional neural networks (CNN) \cite{kalchbrenner2014convolutional}, and tree neural networks \cite{tai2015improved,zhu2015long}.
There are two major differences between these methods: one is the scope in which words can interact with each other. The other is the compositional functions they used.
For example, LSTMs can explicitly model the dependencies between the current word and previous ones, and words can interact with others within the same window. More examples can be found in Table \ref{tab:all-models}.

\paragraph{Attention-based Sequence Modelling.}
Several attention-based mechanisms have been introduced to expand the interaction scope, enabling more words to interact with each together.
For example,
\newcite{yang2016hierarchical,lin2017structured} utilize a learnable query vector to aggregate the weighted information of each word, which can be used to get sentence-level representations for classification tasks.
\newcite{cheng2016long} augment neural networks with a re-reading ability while processing each word.
\newcite{vaswani2017attention} proposes to model the dependencies between words based entirely on self attention without any recurrent or convolutional layers.
The so-called Transformer (SelfAtt3) has achieved state-of-the-art results on a machine translation task.
In contrast with previous work, we introduce different locality biases to learn complex structure of sentences and leverage rich local dependencies within a particular neighbourhood.

In computer vision community, \newcite{wang2017non} also incorporate non-local bias into neural network for image representation.
However, in this work, we focus on sequence learning, which is different from image processing and requires rich contextual information.

\paragraph{Graph Convolutional Neural Network.}
Graph convolutional neural networks (GCNN) have been used to learn the representations of graph structure data.
\newcite{kipf2016semi} proposes a GCNN for semi-supervised graph classification by learning node representations;
\newcite{gilmer2017neural} propose a general neural message passing algorithm to predict the properties of molecule structures.
\newcite{velickovic2017graph} proposes graph attention networks to model graph-structure data, such as protein and citation networks.
While ``\textsc{Graph}'' has been well-studied in above areas, it's still less clear how to build a graph for a sentence.
Besides, our work is different from \cite{velickovic2017graph} lying in the following aspects:
1) We propose to incorporate non-local bias into GNNs while \newcite{velickovic2017graph} aims to combine attention with GNNs.
2) We focus on sentence understanding, which also calls for contextualized word representations.
Some existing works \newcite{marcheggiani2017encoding} try to  construct a graph from a sequence based on pre-defined structures, such as syntactic dependencies, which are not fit to the complexity of sentences' structures since in a specific task, the true dependency structures among the words in a sentence can significantly differ from the given input structures \cite{vaswani2017attention}.
In this paper, we integrate non-local bias into GNNs aim to learn the structures of sentences on the fly based on different tasks.

\section{Contextualized Non-local Neural Networks (CN$^\textbf{3}$)}
To better learn the representation of sequences in terms of contextualized representations of words \cite{peters2018deep} and complex structures \cite{biber1998corpus}, we propose the contextualized non-local network, which incorporates non-local bias into graph neural networks. Generally, CN$^\textbf{3}$ can not only support the encoding of local contextual information, it can also learn task-dependent structures on the fly.

Next, we introduce the contextualized non-local network, which involves two steps. The first is to construct graphs from sentences and introduce contextualized information from attributes. The second is to dynamically update the graph structures for specific tasks. The overall learning process for learning task-dependent graphs is illustrated in Algorithm~\ref{alg:1}.

\subsection{Graph Construction}
The first part of our framework is to construct meaningful graphs from sentences in order to encode different types of local contextual information.
\paragraph{Nodes}
Given a sentence $X = x_1,x_2,\cdots,x_n$, we treat each word $x_i$ as a node.
For each node $i$, we define \textsc{node attributes} ($\mathcal{S}_{v_i} = v^1, \cdots, v^{|\mathcal{S}_{v}|}$) as different types of extra information besides the word identity.
In this paper, we list several typical node attributes described as follows:

\begin{description}
  \item[Position Information] The position of each word in a sentence.
  \item[Contextual Information] Information of nodes can also be enriched by short-term contexts, which can be obtained by convolutional neural network or long-short term memory network. Introducing this information means that we have augmented the model with a local bias: adjacent units are prone to provide useful information \cite{battaglia2018relational,cheng2016long}.
  \item[Tag Information] More prior knowledge such as POS tags can be encoded into the nodes.
\end{description}

\paragraph{Edges}
For each edge between any word pair $(x_i, x_j)$, we define \textsc{edge attributes} ($\mathcal{S}_{e_ij} = e_{ij}^1, \cdots, e_{ij}^{|\mathcal{S}_{e_{ij}}|}$) as combined relationships, such as lexical relationships, syntactic dependency obtained from parsing tree, or co-occurrence relationships.

\paragraph{Contextualized Node Representations}
Each word and attribute will be first converted to be a real-valued vector through a look-up table, and if only the word identity is used, the node representations are simply as word representations. The information of word then will be enriched by fusing its attributes therefore constructing contextualized node representations.

\subsection{Dynamic Graph Structure Updating}
Non-local bias in CN$^\textbf{3}$ is realized as pairwise interactions between any two words, which makes it possible to dynamically learn sentence structures that are geared towards specific tasks.
Specifically, The model first encodes each node in a low-dimensional vector. Afterwards, different nodes communicate with each other via messages, which updates the graph structures and the node representations.
After the nodes are represented by low-dimensional vectors, we create multiple message passing layers $l=1,2, \ldots, L$, which either update the graph structure or the node representations.

\paragraph{Graph Structure Updating}
The graph structures are updated by promoting communication between different nodes. Specifically, let $\mathbf{\alpha}_{ik}$ denote the strength of the relationship between nodes $i$ and $k$. We update  $\mathbf{\alpha}_{ik}$ according to the following formula:
\begin{align}
{s}_{ik} &= f(\bh_k^{l}, \bh_i^{l}, \bv_{i}, \bv_{k}, e_{ik}) \\
         &= \bu^{T} \tanh(\bW[\bh_k^{l}, \bh_i^{l}, \bv_{i}, \bv_{k}, e_{ik}]) \\ \label{eq:att}
\alpha_{ik} &= \softmax({s_{ik}}),
\end{align}
where $\bv$ denotes the node attributes. $e_{i,k}$ represents the edge attributes between node $i$ and $k$, $\bh_i^{l}$ is the representation of node $i$ in layer $l$. $\bu$ and $\bW$ are parameters that can be learned by backpropagation.
$\bh^{0}$ denotes the word embedding.

\begin{algorithm}[t]
\caption{Learning Processes of Contextualized Non-local Neural Networks for Sequences}
\label{alg:metanet}
\begin{algorithmic}[1]
{\footnotesize
\Require Tag sequence $Y = \{y_1,y_2,\ldots,y_T\}$ and text sequence $X = \{x_1,x_2,\ldots, x_T\}$ from a specific task.
\Require A set of node attributes $\mathcal{S}_{v} = v^1, \cdots, v^{|\mathcal{S}_{v}|}$ and edge attributes $\mathcal{S}_{e} = e^1, \cdots, e^{|\mathcal{S}_{e}|}$

\For{$l \in \lbrace 1 \cdots  L \rbrace$}
    \For{$i \in \lbrace 1 \cdots  N \rbrace$} \Comment{\textcolor[rgb]{0.00,0.59,0.00}{Graph Construction}}

        \State  \textcolor[rgb]{0.00,0.59,0.00}{{// \textit{Contextualized representations}}}
        \State{$\mathbf{v_i}$ = Concat-RealValue($\mathcal{S}_{v}$)}
        \For{$j \in \lbrace 1 \cdots  N \rbrace$}
            \State{$\mathbf{e_{ij}}$ = Concat-RealValue($\mathcal{S}_{e}$)}
        \EndFor
    \EndFor
    
    \For{$i \in \lbrace 1 \cdots  N \rbrace$} \Comment{\textcolor[rgb]{0.00,0.59,0.00}{Graph Learning}}
        \For{$j \in \lbrace 1 \cdots  N \rbrace$}

             \State  \textcolor[rgb]{0.00,0.59,0.00}{{// \textit{Dynamic Structure Updating}}}
            \State{${\alpha_{ij}}^l \leftarrow $Edge-Update$(\h_i^l,\h_j^l,\mathbf{v_i},\mathbf{v_j},\mathbf{e_{ij},\theta_e})$}
        \EndFor
        \State{ $\mathbf{h}_i^{l+1} \leftarrow$ Node-Update$(\mathcal{H}^l,\boldsymbol{\alpha_i}^l) $}
    \EndFor

    \Comment{\textcolor[rgb]{0.00,0.59,0.00}{Application Layer}}
    \If{\texttt{Node-Level}}
        \State{$p (Y | X) = \CRF(\h_i^{L},\theta^{(c)})$}
    \ElsIf{\texttt{Graph-Level}}
        \State{$\bh^{L} = \frac{1}{N}\sum_{i}{{\bh}_i^{L}},$}
        \State{$p (Y | X) = \Softmax(\W\bh^{L} + \mathbf{b})$}
    \EndIf

\EndFor

}
  \end{algorithmic} 
  \label{alg:1}
\end{algorithm}

\paragraph{Node Updating}
Once the graph structure $\alpha_{ik}$ is updated, for each node $k$, it first aggregates information from its neighbors. Specifically,
let $\tilde{\bh}_k^{l}$ denote the information collected from neighbors for node $k$ in the $l$-th graph layer, which can be simply calculated as:

\begin{align}
 \tilde{\bh}_k^{l} = \boldsymbol{\alpha_{i}}\mathcal{H}^l =\sum_{i}^{m} \alpha_{ik} \bh_i^{l} \label{eq:alpha}
\end{align}

Afterwards, we update the node representations based on the current node representations $\bh_k^{l}$ and the information aggregated from its neighbors $ \tilde{\bh}_k^{l}$. Here, we choose a gating-based updating methods.

\begin{align}
 {\bh}_k^{l+1} &=  \bg \odot \tilde{\bh}_k^{l} +  (1-\bg) \odot \bh_k^{l}
\end{align}
where $\odot$ denotes element-wise multiplication, and
\begin{align}
\bg &= \sigma(\bW \bh_k^{l} + \bb),
\end{align}
where $\sigma$ represents the logistic function. $\bW$ and $\bb$ are learnable parameters.

\subsection{Application Layers}
After multiple steps of graph structure updating ($l=1,2,\ldots, L$),
we can obtain node-level representations of the final layer $L$ : ${\bh}_1^{L}, ..., {\bh}_m^{L}$, which can be used for different NLP
tasks when followed by different output layers.
Here, we show how to utilize these representations at the node level for POS, Chunking, NER tasks, and graph-level for text classification, semantic matching tasks.

\paragraph{Node-level Representations}
A direct application in this scenario is sequence labelling, which aims to assign a tag sequence $Y = \{y_1,y_2,\ldots,y_T\}$ to a text sequence $X = \{x_1,x_2,\ldots, x_T\}$.
The output at each time step ${\bh}_i^{L}$  can be regarded as the representation of the preceding subsequence, which is then fed into a CRF layer that calculates scores for corresponding tag categories.
Then, in the CRF layer, the conditional probability  $p (Y | X)$ is formalized as:
\begin{align}
    p (Y | X) = \CRF(\h_i^{L},\theta^{(c)})
\end{align}
where $\CRF$ represents the CRF layer and $\theta^{(c)}$ is learnable parameters.
For detailed formulation of the CRF layer, see additional resources.

\paragraph{Graph-level Representations}
In this scenario, we should compute a representation for the whole graph.
The simplest way is to take the average of all the node representations:
$\bh^{L} = \frac{1}{m}\sum_{i}{{\bh}_i^{L}}$


Then, the representation $\bh^{L}$ can be further fed into a softmax function to yield a probability distribution over the output labels for text classification or semantic matching tasks.

\section{Experiments}
\label{sec::experiment}
In this section, we evaluate the effectiveness of our proposed method on ten tasks.
Next, we will give brief descriptions of the tasks and datasets.
For more details of the above dataset descriptions and training hyper-parameters, please see the Additional Resources section.

\subsection{Tasks and Datasets}
We evaluate our models on five text classification tasks, two semantic matching tasks and three sequence labelling tasks.
\begin{itemize}
    \item {\textbf{Text Classification}}: QC (Question Classification); SST2 (the Stanford Sentiment Treebank); MR (The movie reviews with two classes \cite{pang2005seeing}); IMDB (Long text movie reviews.)
    \item {\textbf{Semantic Matching}}: In this task, given two sentences A and B, the model is used to determine the semantic relationship between these two sentences. We choose two typical datasets SICK \cite{marelli2014semeval} and SNLI \cite{bowman-EtAl:2015:EMNLP} for this tasks.
    \item {\textbf{Sequence Labelling}}: We choose POS, Chunking and NER as evaluation tasks on Penn Treebank, CoNLL 2000 and CoNLL 2003 respectively.
\end{itemize}

\begin{table*}[!t]\small
\renewcommand\arraystretch{1}
\center
\begin{tabular}{l*{11}{c}}
\toprule
\multirow{2}{*}{\textbf{Models}}  & \multicolumn{5}{c}{\textbf{Text Classification}} & \multicolumn{2}{c}{\textbf{Semantic Matching}} & \multicolumn{3}{c}{\textbf{Sequence Labelling}}\\
\cmidrule(lr){2-6}  \cmidrule(lr){7-8} \cmidrule(lr){9-11}
 & QC & MR & SST & SUBJ & IMDB & SICK & SNLI & POS & Chunking & NER\\
\midrule
\multicolumn{1}{l}{NBOW}           & 88.2 & 77.2 & 80.5 & 91.3 & 87.5 & 73.4 & 75.1 &  96.38  & 90.51 & 87.91 \\
\midrule
\multicolumn{11}{l}{\textsc{Neural network with Different Structure Bias}} \\
\midrule
\multicolumn{1}{l}{CNN}              & 92.2 & 81.5 & \textbf{88.1} & 93.2 & 88.5  & 75.4 & 77.8  & 97.20 & 93.63 & 88.67 \\
\multicolumn{1}{l}{RNN}              & 90.2 & 77.2 & 85.0 & 92.1 & 87.0 & 74.9  & 76.8 & 97.30 & 92.56 & 88.50\\
\multicolumn{1}{l}{LSTM}             & 91.3 & 77.7 & 85.8 & 92.5 & 88.0 & 76.3  & 77.6 & 97.55 & 94.46 & 90.10\\
\multicolumn{1}{l}{TreeLSTM}         & 92.8 & 78.7 & 88.0 & 93.3 &  $\times$    & 77.5  & 78.3  & - & - & -\\
\multicolumn{1}{l}{PDGraph}          & 93.2 & 80.2 & 86.8 & 92.5 & $\times$     & 78.8  & 78.9  & - & - & -\\
\midrule
\multicolumn{11}{l}{\textsc{Attention-based Models}} \\
\midrule
\multicolumn{1}{l}{SelfAtt1}          & 93.2 & 80.3 & 86.4 & 92.9 & 90.3 & 78.4 & 77.4 & * & * & * \\
\multicolumn{1}{l}{SelfAtt2}          & 93.5 & 79.8 & 87.0 & 93.1 & 89.2 & 79.8 & 79.2  & - & - & - \\
\multicolumn{1}{l}{SelfAtt3}          & 90.1 & 77.4 & 83.6 & 92.2 & 88.5 & 76.3 & 76.9 & 96.42  & 91.12 & 87.61\\
\midrule
\multicolumn{11}{l}{\textsc{Task-Dependent Graphs}} \\
\midrule
\multicolumn{1}{l}{CN$^\textbf{3}$$_{LSTM}$}                 & 94.2 & 81.3 & 87.5 & 93.5  & 91.5  & 80.2   & 81.3 & 97.12 & 93.81 & 89.33\\ [0.04cm]
\multicolumn{1}{l}{CN$^\textbf{3}$$_{LSTM+POS}$}             & 94.8 & 80.7 & 87.1 & 94.3  & 91.0  & 81.4   & 82.1 & - & - & -\\ [0.04cm]
\multicolumn{1}{l}{CN$^\textbf{3}$$_{LSTM+char}$}            & 94.6 & \textbf{82.5} & {88.0} & 94.0  & \textbf{92.5}  & 81.5   & 82.7 & 97.65 & 94.82 & 90.51\\ [0.05cm]
\multicolumn{1}{l}{CN$^\textbf{3}$$_{LSTM}^{Dep}$}           & \textbf{95.0} & 81.0 & 87.8 & \textbf{94.6}  & *     & \textbf{81.9}   & \textbf{83.5} & -     & -     & -\\
\multicolumn{1}{l}{CN$^\textbf{3}$$_{LSTM+char+Spell}$}      & -    & -    & -    & -     & -     & -      & -    & \textbf{97.78} & \textbf{95.13} & \textbf{91.10}\\

\midrule
\bottomrule
\end{tabular}
\caption{
Performance of the proposed models on all datasets compared to typical baselines.
$\times$ indicates that corresponding models can not work since the sentences are too long to be processed by parser.
$*$ denotes corresponding models can not be used for sequence labelling tasks.
$-$ denotes corresponding publications don't evaluate models on related tasks.
The superscript of CN$^\textbf{3}$ denotes the edge attributes while subscript represents node attributes. Specifically,
$LSTM$ denotes the information of each node is enriched by long-short-term memory unit. And the $Spell$ feature is an indicator that the first letter of a word is capital or small.
$Dep$ provides the information that if two node have an edge in syntactic dependency tree.  $POS$ denotes the POS taging tags while $char$ represents character information.
\textbf{NBOW}: Sums up the word vectors for graph-level representation and  concatenate word with positional embedding as node-level representation.
\textbf{CNN}: We use \protect \cite{kim2014convolutional} for graph-level representation and \protect \cite{collobert2011natural} for node-level representation.
\textbf{LSTM} We use \protect \cite{tai2015improved} for graph-level representation and \protect \cite{huang2015bidirectional} for node-level representation.
\textbf{TreeLSTM}: LSTM over tree-structure \cite{tai2015improved}.
\textbf{PDGraph}: Pre-defined Graph Neural network based on syntactic dependency. \cite{marcheggiani2017encoding}.
\textbf{SelfAtt1}: A structured self-attentive sentence \protect \cite{lin2017structured}.
\textbf{SelfAtt2}: Proposed by \protect \cite{cheng2016long}.
\textbf{SelfAtt3}: Also known as Transformer, which is proposed by \protect \cite{vaswani2017attention}.
}\label{tab:10tasks}
\end{table*}

We parse the sentences in the datasets with Stanford NLP toolkit \cite{manning2014stanford} to obtain dependency relations and Part-of-Speech tags for our models and several competitor models.

\subsection{Settings}
To minimize the objective, we use stochastic gradient descent with the diagonal variant of AdaDelta \cite{zeiler2012adadelta}.
The word embeddings for all of the models are initialized with GloVe vectors \cite{pennington2014glove}.
The other parameters are initialized by randomly sampling from a uniform distribution in $[-0.1, 0.1]$.
For each task, we take the hyperparameters which achieve the best performance on the development set via grid search.
Other detailed settings of our models can be seen in the Additional Resources section.

\subsection{Quantitative Evaluation}
The proposed models support highly flexible graph representations in two ways: first, in terms of the representation of the attributes (feature encoding of nodes and edges); and second, in terms of the structure of the graph itself (dynamic learning of task-dependent structure). Next, we will elaborate on these evaluations.

\paragraph{Evaluation on Dynamic Structure Learning}
Table~\ref{tab:10tasks} shows the performances of different models on 10 different tasks. We have  following observations:
\begin{itemize}
    \item For graph-level task, CN$^\textbf{3}$$_{LSTM}$ consistently outperforms neural networks with different structural biases (sequential, tree, pre-defined graph) and attention mechanisms, indicating the effectiveness of non-local bias and dynamic learning nature for sentence structures.
    Particularly, compared with PDGraph, CN$^\textbf{3}$$_{LSTM}$ achieves better performance and doesn't rely on external syntactic tree, with the ability to handle more longer texts, such as IMDB.
    Compared with LSTM, the improvement of CN$^\textbf{3}$$_{LSTM}$ also indicates the functions of local and non-local biases are complementary.
    \item For node-level tasks (sequence labelling), CN$^\textbf{3}$$_{LSTM}$ achieves comparable results as opposed to CNN (which utilizes mulit-task learning framework) and LSTM (which additionally introduces many external features: gazetteer features and spelling features).
\end{itemize}

\paragraph{Evaluation on Attributes}
The proposed model has the advantage of being able to contextualize words by encoding information of nodes' or edges' attributes. Here we use superscripts and subscripts to introduce nodes' or edges' attributes.
Table \ref{tab:10tasks} illustrates:

\begin{itemize}
    \item Compared to SelfAtt3 (Transformer), CN$^\textbf{3}$$_{LSTM}$ obtained substantial improvements, especially on SST dataset constructed by lots of sentences with complicated sentence patterns.  For those node-level tasks, CN$^\textbf{3}$$_{LSTM}$ surpasses SelfAtt3 and we attribute the success to its power in both the ability of learning structures dynamically and encoding short-term contextual information.
    \item The performances can be enhanced when rich node or edge attributes were taken into accounts. And we observed that effects of different attributions are different. Specifically, CN$^\textbf{3}$$_{LSTM+char}$ achieves best performances on MR and IMDB datasets while CN$^\textbf{3}$$_{LSTM}^{Dep}$ has obtained best performances on QC, SUBJ, SICK and SNLI datasets. The reason is that the texts in QC, SUBJ, SICK and SNLI is more formal where higher accuracies on external tools (Parser or tagger) was achieved, while for MR, and IMDB, they contain more informal expressions such as ```\texttt{cooool, gooood}''', leading to the failure of external linguistic tools  though they can be resolved by character-aware models.
    \item While introducing the same spelling features with LSTM model, the performances of CN$^\textbf{3}$$_{LSTM+char+Spell}$ in tagging tasks can be further improved.
    To make an complete comparison, we have also listed the performances of our models in tagging tasks aginst state-of-the art models in Tab.\ref{tab:exp-st}\footnote{Since it's hard to integrate this table into Tab.\ref{tab:10tasks} }. Competitor models in Tab.\ref{tab:exp-st} or introduce multi-task learning methods \cite{collobert2008unified,yang2016multi}, or utilize more unsupervised knowledge \cite{peters2018deep,yasunaga2018robust}, or design more handcrafted features \cite{ma2016end}. Rather, proposed models utilize less knowledge in terms of data or external features while achieving comparable results.
\end{itemize}

\begin{table}[!t]
\small
\centering
\begin{tabular}{p{3.0cm}ccc}
\toprule
\textbf{Model} & Chunking & NER & POS \\
\midrule
\newcite{collobert2008unified}          & 94.32  & 89.59 & 97.29 \\
\newcite{yang2016multi}          & \textbf{95.41}  & 90.94 & 97.55 \\
\newcite{peters2018deep} & -  & \textbf{92.22} & -\\
\newcite{yasunaga2018robust} & - & - & 97.58 \\
\newcite{ma2016end} & - & 91.21 & 97.55 \\

\midrule
Ours        & 95.13  & 91.10 & \textbf{97.78} \\
\bottomrule
\end{tabular}
\caption{Performances of our model against state-of-the-art models.
} \label{tab:exp-st}
\end{table}

\begin{table*}[!t]
\small

\centering
\begin{tabular}{llllll}
\toprule
  & \multicolumn{4}{c}{\textbf{Interpretable Sub-Structures}} & \textbf{Explanations}\\
\midrule
\textbf{Semantic}
& \raisebox{-.5\height}{\includegraphics[width=0.11\textwidth]{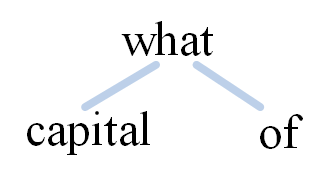}}
& \raisebox{-.5\height}{\includegraphics[width=0.12\textwidth]{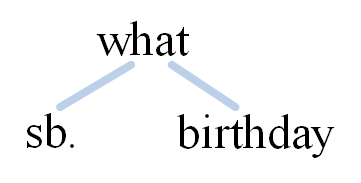}}
& \raisebox{-.5\height}{\includegraphics[width=0.11\textwidth]{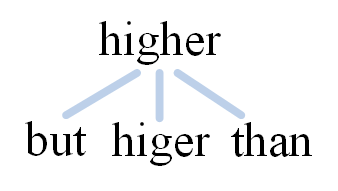}}
& \raisebox{-.5\height}{\includegraphics[width=0.11\textwidth]{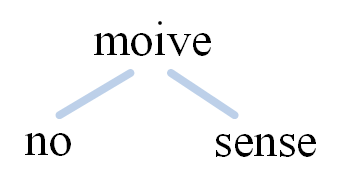}}
& \tabincell{l}{Key sentence patterns for question  \\classification and sentiment analysis \\tasks}
\\
\midrule
\textbf{Syntactic}
& \raisebox{-.5\height}{\includegraphics[width=0.12\textwidth]{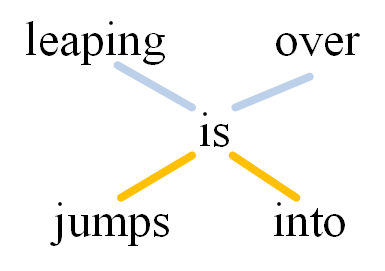}}
& \raisebox{-.5\height}{\includegraphics[width=0.13\textwidth]{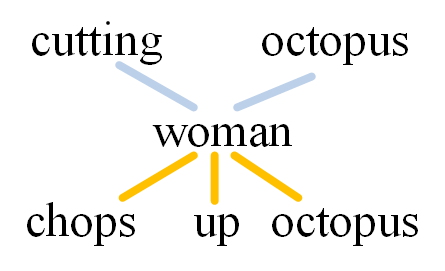}}
& \raisebox{-.5\height}{\includegraphics[width=0.13\textwidth]{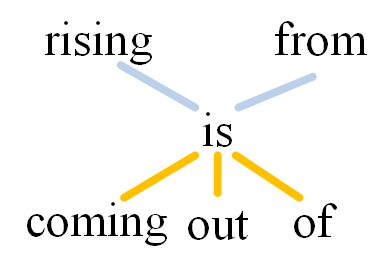}}
& \raisebox{-.5\height}{\includegraphics[width=0.12\textwidth]{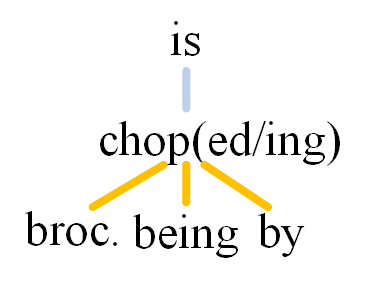}}
& \tabincell{l}{Key Subgraph pairs for semantic \\ matching task. Relating to different\\ types of phrases (verb-adverb, verb-\\object) and voices (active or passive)}
\\
\bottomrule
\end{tabular}
\caption{Multiple interpretable sub-structures generated by different tasks.
For text classification tasks, we show the sub-structure of a sentence, which is a key pattern for final prediction.
For semantic matching, since there are two sentences as input, we simultaneously give a pair of sub-structures with different color lines.
For example, the subgraph pair in the second row and column denote the structure of ``\texttt{is leaping over}'' is learned in the first sentence and simultaneously
``\texttt{is jumping into}'' is learned in the second sentence.
} \label{tab:ed-1}
\end{table*}

\subsection{Qualitative Analysis}

In this section, we aim to better understand our models in terms of the following two aspects:
1) How do learned task-dependent structures contribute to different tasks?
2) How do these attributes influence the learned structures?
We design a series of experiments to address these questions.

\paragraph{Analysis on Task-dependent Structures}
Since the proposed models can explicitly learn the relationship between each word pair by computing edge weights, we randomly sample several examples across different tasks (text classification and semantic matching) and visualize their learned structures.

As shown in Table~\ref{tab:ed-1}, there are multiple interpretable sub-structures. We observe the following points:
\begin{itemize}
    \item For several simple tasks such as text classification, the words in a sentence are usually organized so as to accurately express certain semantic information.

      For example, for a question classification task, the informative patterns constructed by the question word ``\texttt{what}'' can be easily learned.
      For a sentiment classification task, the word ``\texttt{movie}'' is prone to connecting to sentimental words, such as  ``\texttt{terrible}'', ``\texttt{no sense}''.

\item For more complex tasks such as semantic matching, a ground-truth understanding of the syntactic structure is important. In this context, we find that, given a sentence pair, our model is more likely to learn their syntactic information.
    For example, for the sentence pair ``\texttt{A man is rising from a swamp/A man is coming out of the water}'',	our model learns that ``\texttt{is rising from}'' and ``\texttt{is coming out of}'' are two informative patterns in two sentences respectively, which are crucial for accurately predicting the relationship of the sentence pair (``\textit{Entailment}'').
\end{itemize}

\paragraph{Analysis on Attributes}
As the above results demonstrate, attributes (Part-of-Speech or dependency relation) will influence the learning process of relationships among different nodes.
In order to obtain a better intuitive understanding, we randomly pick samples from the dataset (QC), and compare the learned dependencies among words learned by CN$^\textbf{3}$$_{LSTM}$ and CN$^\textbf{3}$$_{LSTM}^{Dep}$.

\begin{figure}[t]
\setlength{\belowcaptionskip}{-0.1cm}
\centering
 \includegraphics[width=0.70\linewidth]{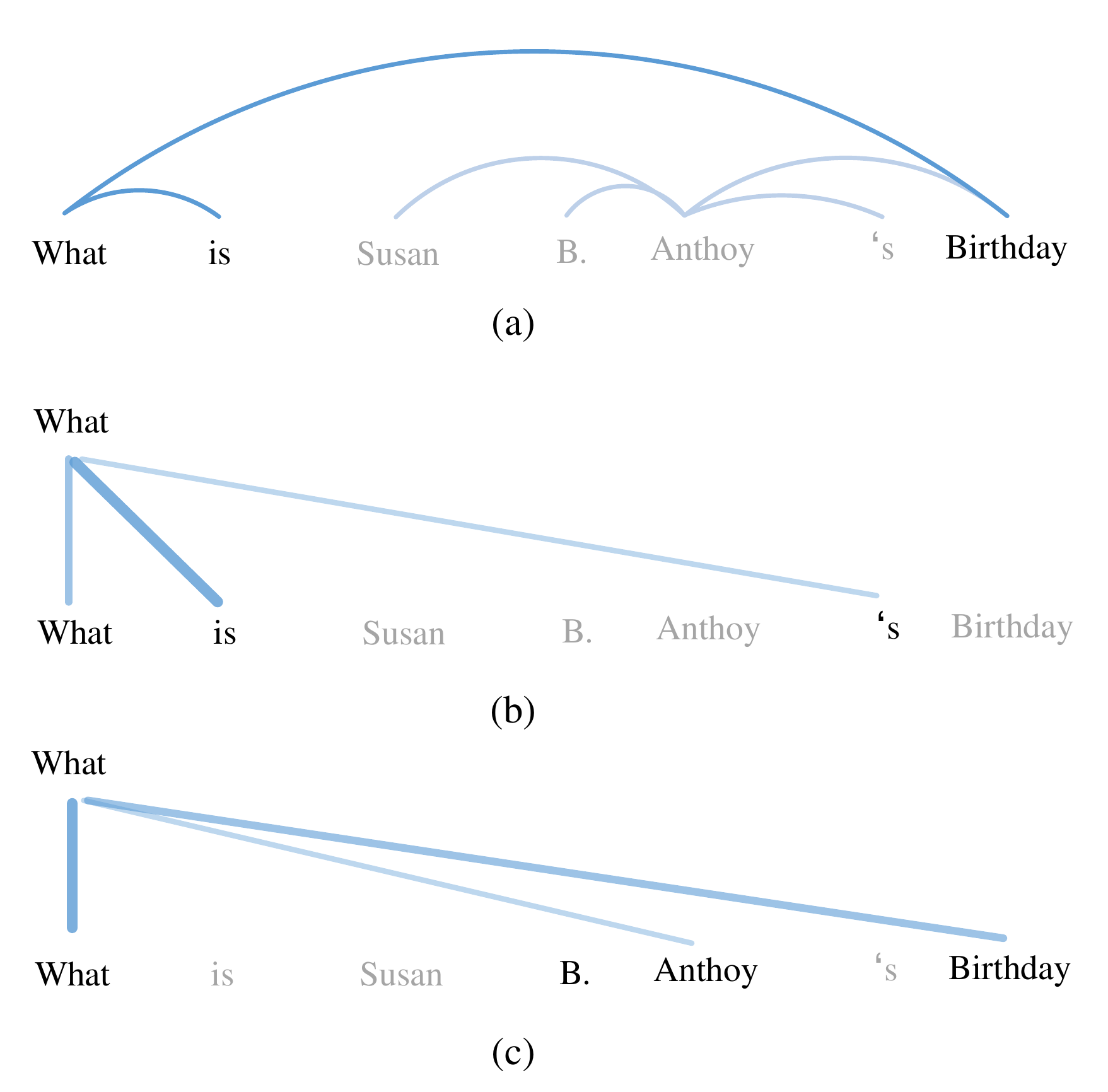}
 \caption{
 (a) The dependency tree obtained by the Stanford Parser.
 (b-c) The relationship between ``\texttt{What}'' and other words, which are learned by CN$^\textbf{3}$$_{LSTM}$ and CN$^\textbf{3}$$_{LSTM}^{Dep}$ models respectively. The correct label of ``\texttt{What is Susan B. Anthoy 's birthday ?}'' is ``\texttt{Number}'', indicating that it's a question asked about ``\texttt{Number}''.
 }\label{fig:exp-num}
\end{figure}

As shown in Figure~\ref{fig:exp-num} (b-c), given a sentence, CN$^\textbf{3}$$_{LSTM}^{Dep}$ makes a correct prediction about the type of this question while CN$^\textbf{3}$$_{LSTM}$ fails. We note the following points.

1) With the help of edge attributes, more useful patterns can be built into the graph structure. For example, ``\texttt{What}'' strongly connects to the words ``\texttt{What}'' and ``\texttt{birthday}'' therefore it captures the latent sentence pattern ``\texttt{What ... birthday }'', which is the key to predicting the type of this question.

2) {The relationships between different words are task-specific rather than exactly matching the pre-defined syntactic dependencies}. Although our model gets a hint that there is strong connection between ``\texttt{What}'' and ``\texttt{is}'' from the dependency parser as shown in Figure \ref{fig:exp-num} (a), yet CN$^\textbf{3}$$_{LSTM}^{Dep}$ regards it as less informative. We think it simply reflects the

\section{Conclusion}
In this paper, we first analyze two challenges for sequence learning from the perspective of a model's locality biases, which motivates us to examine the complementary natures of Transformer and graph neural networks. Then, we draw on their complementary strengths to propose a contextualized non-local neural network.
Experimental results show that learning task-dependent structures of sentences and contextualized word representations are crucial to many NLP tasks.

\section*{Acknowledgments}
The authors wish to thank the anonymous reviewers for their helpful comments. This work was partially funded by National Natural Science Foundation of China (No. 61751201, 61672162), STCSM (No.16JC1420401, No.17JC1404100), and Natural Sciences and Engineering Research Council of
Canada (NSERC).

\bibliographystyle{named}
\bibliography{./fig/nlp1,./fig/ours1,./fig/nlp,./fig/ours2}

\end{document}



\section{Additional Resources}

\subsection{CRF Layer}
More formally, In CRF layer, $p (Y | X)$ is formalized as:
\begin{equation}
p (Y | X) = \frac{\Psi (Y | X)}{\sum_{Y^\prime \in \L^n} \Psi (Y^\prime | X)}.
\end{equation}
Here, $\Psi (Y | X)$ is the potential function, and we only consider interactions between two successive labels (first order linear chain CRFs):
\begin{gather}
\Psi (Y | X) = \prod_{i = 2}^n \psi (X, i, y_{i-1}, y_i),\\
\psi (\x, i, y^\prime, y) = \exp(s(X, i)_{y} + \bb_{y^\prime y}),
\end{gather}
where $\bb_{y^\prime y} \in \R$ is trainable parameters respective to label pair $(y^\prime, y)$. Score function $s(X, i) \in \mathbb{R}^{|\L|}$ assigns score for each label on tagging the $i$-th word:
\begin{equation}
s(X, i) = \bW^\top {\bh}_i^{L} + \bb,
\end{equation}
where ${\bh}_i^{L}$ is the hidden state of $L$-th graph layer at position $i$; $\bW \in \mathbb{R}^{d_h \times |\mathcal{L}|}$ and $\bb \in \mathbb{R}^{|\mathcal{L}|}$ are trainable parameters.

\subsection{Tasks and Datasets}

\paragraph{Text Classification} In this task, we evaluate our models on five different datasets.
\begin{itemize*}
  \item \textbf{QC} The TREC questions dataset involves six different question types. \cite{li2002learning}.
  \item \textbf{SST2} The movie reviews with two classes (negative, positive) in the Stanford Sentiment Treebank \cite{socher2013recursive}. 
 \item \textbf{MR} The movie reviews with two classes \cite{pang2005seeing}. 
 \item \textbf{SUBJ} Subjectivity dataset where the goal is to classify each instance (snippet) as being subjective or objective. \cite{pang2004sentimental}
 \item \textbf{IMDB} The IMDB dataset consists of more longer movie reviews with binary classes.
\end{itemize*}

\paragraph{Semantic Matching}


In this task, given two sentences A and B, the model is used to determine the semantic relationship between these two sentences.
We choose two typical datasets for this tasks.
\begin{itemize*}
    \item \textbf{SICK} This dataset is proposed by \cite{marelli2014semeval} and consists of 9927 sentence pairs in a 4500/500/4927 train/dev/test split.
    \item \textbf{SNLI}  We use the Stanford Natural Language Inference Corpus (SNLI) \cite{bowman-EtAl:2015:EMNLP}, which has been a standard benchmark for semantic matching.
\end{itemize*}

\paragraph{Sequence Labelling}
For sequence labelling task, we choose POS, Chunking and NER as evaluation tasks.
\begin{itemize*}
    \item \textbf{Penn Treebank} Sections 0-18 of Wall Street Journal (WSJ) data are used for training, while sections 19-21 are for validation and sections 22-24 for testing.
    \item \textbf{CoNLL 2000} Chunking is evaluated using the CoNLL 2000 shared task.
    \item \textbf{CoNLL 2003} NER is evaluated using the CoNLL 2003 shared task.
\end{itemize*}

In this task, we evaluate our models on five different datasets.
The detailed statistics about the five datasets are listed in Table \ref{tab:data}.
Each dataset is briefly described as follows.

\begin{table}[!t]\small\centering
\begin{tabular}{*{7}{c}}

 \toprule
 \textbf{Dataset} & Train & Dev. & Test & Class & $L_{avg}$ & $|\mathcal{V}|$\\
 \midrule
 MR    & 9596  & -     & 1066 & 2 & 22 & 21K\\
 SST2   & 6920  & 872   & 1821 & 2 & 18 & 15K\\
 SUBJ  & 9000  & -     & 1000 & 2 & 21 & 21K\\
 QC    &  5452 & -     & 500 & 6 & 10 & 9.4K\\
 IMDB & 12000  & -     & 3000 & 2 & 198 & 112K \\
 \bottomrule
\end{tabular}
\caption{Statistics of the five mainstream datasets for text classification. $L_{avg}$
denotes the average length of documents; $|\mathcal{V}|$ denotes the size of vocabulary.}\label{tab:data}
\end{table}

\begin{table}[!t]\small
\center
\tabcolsep0.06in
\begin{tabular}{l*{6}{l}}
\toprule
\textbf{Dataset} &       \textbf{Task} &	 	 \textbf{Training } &	 \textbf{Dev. } &	 \textbf{Test } \\
\midrule

CoNLL 2000 &\multirow{1}{*}{Chunking}
&  211,727 &  - &  47,377    \\
\midrule
CoNLL 2003 &\multirow{1}{*}{NER}
&  204,567 &  51,578 &  46,666    \\
\midrule
PTB &  \multirow{1}{*}{POS}
&  912,344 &  131,768 &  129,654     \\

\bottomrule
\end{tabular}
\caption{The sizes of the sequence labelling datasets in our experiments, in terms of the number of tokens.}\label{tab:st}
\end{table}

\begin{table*}[!t]\small\centering
\begin{tabular}{*{11}{c}}

 \toprule
 \textbf{Dataset} & QC & MR & SST & SUBJ & IMDB & SICK & SNLI & POS & Chunking & NER\\
 \midrule
 $d$(word)    &  100  & 200   & 200  & 100 & 200 & 50 & 300  & 100 & 100 & 100 \\
 $d$(char)    &  50   & 50    & 50   & 50  & 50  & 20 & 50   & 50  & 50  & 50\\
 $d$(pos)     &  20   & 20    & 20   & 20  & 20  & 20 & 50   & 20  & 20  & 20\\
 $d$(POS)     &  20   & 20    & 20   & 20  & 20  & 20 & 50   & -   & -   & -\\
 $d$(Spell)   &  -    & -     & -    & -   & -   & -  & -    & 10  & 10  & 10\\
 \bottomrule
\end{tabular}
\caption{Statistics of hyper-parameters of CN$^\textbf{3}$. $d$(word) denotes the size of word embedding.   ``pos'' and ``POS'' represent positional information and Part-of-Speech tag.}\label{tab:hyper}
\end{table*}

\subsection{Training and Hyperparameters}

%
To minimize the objective, we use stochastic gradient descent with the diagonal variant of AdaDelta \cite{zeiler2012adadelta}.
The word embeddings for all of the models are initialized with GloVe vectors \cite{pennington2014glove}.
The other parameters are initialized by randomly sampling from uniform distribution in $[-0.1, 0.1]$.
For each task, we take the hyperparameters which achieve the best performance on the development set via a small grid search.
Other detailed settings of our models are listed in Tab.\ref{tab:hyper}.

\subsection{Evaluation on Multiple-step Learning}
We also investigate the effect of multiple graph propagation on text classfication tasks.
\paragraph{Quantitative Analysis}
We find that the performance can be improved with the help of increasing graph layers.
As shown in  Figure \ref{fig:layers}, the performance of two typical models (CN$^\textbf{3}$$_{pos}$ and CN$^\textbf{3}$$_{LSTM}$) changes over different layers.
Specifically, for CN$^\textbf{3}$$_{LSTM}$, two layers are enough to these five datasets, while for CN$^\textbf{3}$$_{pos}$, we can see the best performances are achieved with the help of more layers.
We think the reason is that CN$^\textbf{3}$$_{pos}$  can not model local interaction well, while the multiple graph propagation make up for it.

\begin{figure*}[!t]
\centering
\subfigure[QC]{
 \includegraphics[width=0.18\linewidth]{./standalone/z-QC}\label{fig:groups1}
 }
\subfigure[MR]{
 \includegraphics[width=0.18\linewidth]{./standalone/z-MR}\label{fig:groups1}
 }
\subfigure[SST2]{
 \includegraphics[width=0.18\linewidth]{./standalone/z-SST2}\label{fig:groups1}
 }
\subfigure[SUBJ]{
 \includegraphics[width=0.18\linewidth]{./standalone/z-SUBJ}\label{fig:groups1}
 }
\subfigure[IMDB]{
 \includegraphics[width=0.18\linewidth]{./standalone/z-IMDB}\label{fig:groups1}
 }
 \caption{Performances of CN$^\textbf{3}$$_{LSTM}$ (green line) and CN$^\textbf{3}$$_{pos}$ (black line) with different numbers of graph layers on five development datasets: QC, MR, SST2, SUBJ, IMDB. Y-axis represents the error rate(\%), and X-axis represents the number of layers.
 }\label{fig:layers}
\end{figure*}

\paragraph{Qualitative Analysis}
The proposed model can learn the relationships between different words by multiple propagation steps, similar to the setting of different layers.
Here, we randomly samples some examples in QC dataset and analyze the learned structures by CN$^\textbf{3}$$_{LSTM}$ under two different layers.

\begin{figure}[!t]
\setlength{\belowcaptionskip}{-0.1cm}
\centering
\includegraphics[width=0.8\linewidth]{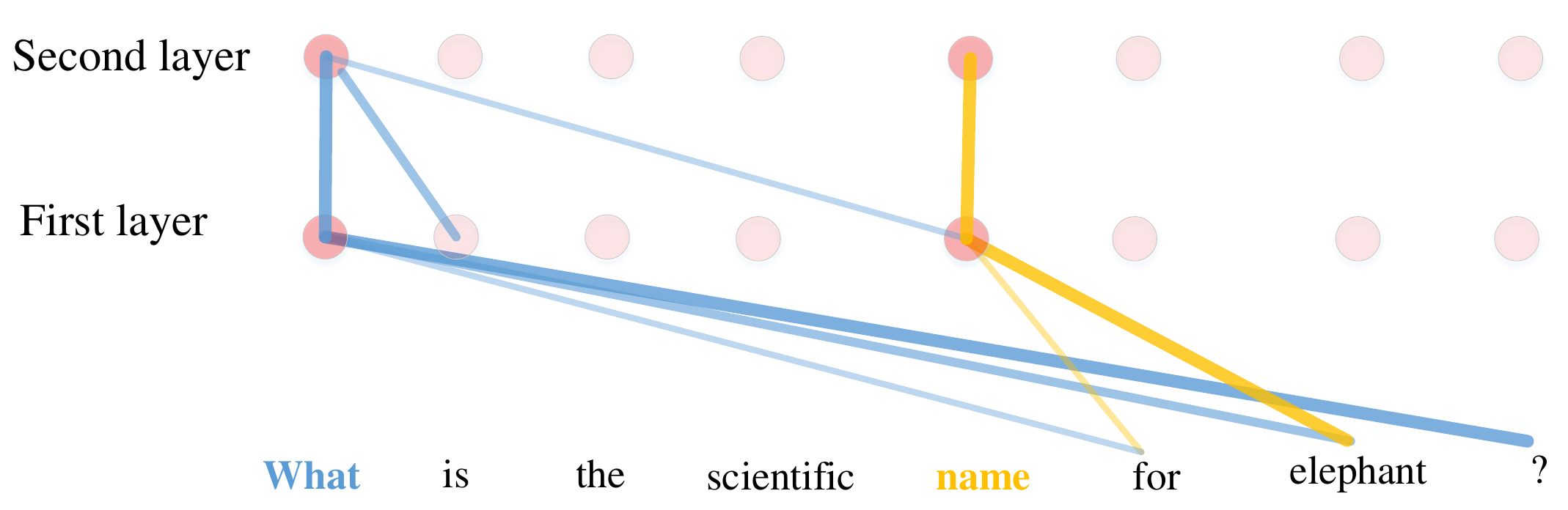}
\caption{
 Illustrations of the learned structures under different layers for the sentence ``\texttt{What is the scientific name of elephant ?}'' . The boldness of the line indicates the strength of the dependency. And we choose two typical words ``\texttt{What}'' and ``\texttt{name}'' to visualize with different colors
}\label{fig:exp-2layer}
\end{figure}

Figure \ref{fig:exp-2layer} illustrates that nodes ``\texttt{What}'' and ``\texttt{name}'' collect different parts of information in different layers.
In the first layer, the word ``\texttt{name}'' gathers the information from ``\texttt{for elephant}'', and then in the second layer,
the node ``\texttt{What}'' successfully builds the relationship with the first layer node ``\texttt{name}'', indicating the model has noticed that, to accurately predict the type of the question (Entity), it's crucial to combine the meanings of ``\texttt{What}'' and ``\texttt{name}'' together.

\bibliographystyle{named}
\bibliography{./fig/nlp1,./fig/ours1,./fig/nlp,./fig/ours2}